%% file: hcgl.tex
\documentclass[lettersize,journal]{IEEEtran}
\usepackage{amsmath,amsfonts}
\usepackage{algorithmic}
\usepackage{algorithm}
\usepackage{array}
\usepackage{subcaption}
\usepackage{textcomp}
\usepackage{stfloats}
\usepackage{url}
\usepackage{verbatim}
\usepackage{graphicx}
\usepackage{xcolor}         % colors

\usepackage[square,sort,comma,numbers]{natbib}
\usepackage{booktabs}
\usepackage{bm} 
\usepackage[T1]{fontenc}
\usepackage{newfloat}
\usepackage{listings}

\begin{document}

% Hierarchical Cooperation Graph Learning HCGL
\title{Self-Clustering Hierarchical Multi-Agent Reinforcement Learning with Extensible Cooperation Graph}
\author {
    Qingxu Fu     ,
    Tenghai Qiu   , 
    Jianqiang Yi  , 
    Zhiqiang Pu   , 
    Xiaolin Ai
    \thanks{This work was supported in part by 
    the National Key Research and Development Program of China (2018AAA0102404), 
    the National Natural Science Foundation of China (62073323), 
    the Strategic Priority Research Program of Chinese Academy of Sciences (XDA27030403),
    the External Cooperation Key Project of Chinese Academy Sciences (173211KYSB20200002),
    the Science and Technology Development Fund of Macau (No.0025/2019/AKP),
    and 
    the Beijing Nova Program under Grant 20220484077.
    
    The authors are with the Institute of Automation, Chinese Academy of Sciences, Beijing 100190, China, 
    the School of Artificial Intelligence, University of Chinese Academy of Sciences, Beijing 100049, China,
    and also with CETC Information Science Academy, Beijing 100049, China.
    (e-mail: fuqingxu2019@ia.ac.cn, tenghai.qiu@ia.ac.cn, jianqiang.yi@ia.ac.cn, 
    zhiqiang.pu@ia.ac.cn, xiaolin.ai@ia.ac.cn, Yuanwanmai7@163.com).
    }% <-this % stops a space
}

% The paper headers
\markboth{Journal of \LaTeX\ Class Files,~Vol.~14, No.~8, August~2021}%
{Shell \MakeLowercase{\textit{et al.}}: A Sample Article Using IEEEtran.cls for IEEE Journals}

\maketitle

\begin{abstract}
    \input{tex/01-abstract.tex}
\end{abstract}

\begin{IEEEkeywords}
Multi-agent system, reinforcement learning, hierarchical MARL.
\end{IEEEkeywords}

\input{tex/02-intro.tex}
\input{tex/03-related.tex}
\input{tex/04-method.tex}

\input{tex/05-experiment.tex}
\input{tex/06-conclusion.tex}

\bibliography{citelib.bib}
\bibliographystyle{IEEEtran}

\end{document}

%% file: tex/01-abstract.tex
Multi-agent Reinforcement Learning (MARL) has been successful in solving numerous cooperative challenges.	
However,
classic non-hierarchical MARL algorithms still cannot address various complex multi-agent problems that require hierarchical cooperative behaviors.
The cooperative knowledge and policies learned in non-hierarchical algorithms are implicit and not interpretable, thereby restricting the incorporation of existing knowledge.
This paper proposes a novel hierarchical MARL model called Hierarchical Cooperation Graph Learning (HCGL) that aims to solve general multi-agent problems.
HCGL features a dynamic graph structure named Extensible Cooperation Graph (ECG), which is essential for achieving agents' self-clustering behaviors and hierarchical cooperation.	
ECG comprises three hierarchical layers consisting of agent nodes, cluster nodes and target nodes, respectively.
The primary distinction of our ECG model is that the behavior of agents is directed by the topology of ECG, rather than neural policy networks.
In this paper, we propose a distinctive approach to manipulate the ECG topology in response to changing environmental conditions.
Specifically, we introduce four virtual operators trained to modify the edge connections of the ECG,
and thereby grouping agents into clusters and achieving higher-level cooperative behaviors.
The hierarchical feature of ECG provides a unique approach to merge raw agent actions (executed by individual agents) and cooperative actions (executed by agent clusters) into one unified action space,
allowing us to incorporate fundamental cooperative knowledge into an extensible interface. 
In our experiments, the HCGL model has shown outstanding performance in multi-agent benchmarks with sparse rewards.
We also verify that HCGL can easily be transferred to large-scale scenarios with high zero-shot transfer success rates.

%% file: tex/02-intro.tex
\section{Introduction}

% Reinforcement Learning
Multi-Agent Reinforcement Learning (MARL) has become increasingly important in establishing cooperative agent strategies within large-scale multi-agent systems.
By leveraging deep neural networks,
MARL algorithms can extract vital features from large observation spaces
and perform sophisticated cooperative behaviors.
However, the performance of MARL algorithms is limited by 
various factors existing in the MARL learning frameworks,
particularly when there are a large number of agents \cite{fu2022concentration}.
Exploring feasible cooperation strategies under these circumstances is difficult and computationally expensive, 
even if some of these strategies seem basic or obvious to humans.

% When there are a vast number of agents present,
% it is difficult and computationally expensive to explore feasible cooperation strategies,
% even if some of these strategies seem basic and obvious to us as humans.

% 先前的很多研究探索了内在奖励和奖励分解的应用。奖励分解在智能体数量极多的情况下表现不佳，
% 内在奖励的设计方法很多但很难放在同一框架内，且容易失误引入人类对问题的错误偏见，我们认为应当另辟蹊径
Prior studies have explored the application of intrinsic reward \cite{jaques2019social}, reward decomposition \cite{rashid2018qmix}, etc.
However, reward decomposition shows state-of-the-art performance in small-scale multi-agent tasks, but it performs poorly in scenarios with a large number of agents.
On the other hand, 
the intrinsic reward methods can effectively promote the coverage speed of cooperative behaviors.
Nevertheless,
the intrinsic reward may introduce biased cognition on tasks from reward designers,
leading the agents to learn unexpected behaviors that may not align with the initial objectives.
Moreover, reward-shaping techniques that work well for one problem may turn out to be ineffective for another
and may even result in reward poisoning \cite{rakhsha2021reward}.

Learning high-level cooperative behaviors in large-scale multi-agent environments is still an open question with the following problems:
\begin{enumerate}
    \item Large-scale multi-agent policy exploration problem:
    As the joint action space expands dramatically with the number of agents,
    it grows extremely difficult to explore enough feasible policies for agents
    to escape from local suboptimal solutions.
    This problem can be even more severe in environments where rewards are sparse.
    \item Knowledge incorporation problem:
    Only a limited number of studies investigate blending knowledge into MARL algorithms.
    A framework that enables knowledge integration is critical for two significant reasons. 
    Firstly, 
    integrating data and models we already possess can significantly enhance RL performance.
    Secondly, knowledge integration is necessary to learn cooperative behaviors beyond the limitations of RL.	
    Some cooperative behaviors are established based on environmental factors 
    that cannot be easily simulated. 
    For example, 
    birds' formation behavior is based on aerodynamic factors 
    that help them save energy during long flights. 
    Without expensive aerodynamics calculations,
    RL agents cannot learn even the simplest flight formation as nothing drives them to do so.
    \item Interpretability problem:
    Agents in non-hierarchical frameworks learn advanced high-level policies with neural networks,
    which cannot be easily interpreted. 
    If agents fail to exhibit expected cooperative behaviors after training, 
    diagnosing problems within the agents' policy neural networks is almost impossible.
  \end{enumerate}

This paper proposes Hierarchical Cooperation Graph Learning (HCGL) to address these problems:
\begin{enumerate}
  \item Firstly,
the joint policy exploration problem can be addressed by introducing a general hierarchical control model.
In this paper, HCGL introduces the Extensible Cooperation Graph (ECG), a unique control structure that aims to enable hierarchical multi-agent cooperation while minimizing ineffective exploration behaviors.	
Different from classic MARL algorithms,
agents in HCGL are directly controlled by ECG instead of neural network policies,
their action depends solely on the topology of ECG.
ECG represents agents as nodes and groups agent nodes into several clusters.
Then, each cluster is connected to a series of target nodes 
defined by either \textit{cooperative actions} (cluster-level actions) or \textit{primitive actions} (agent-level actions).
The cooperative actions are high-level actions that are executed by all member agents in a cluster.

\item Secondly,
HCGL employs cooperative actions to blend cooperative knowledge into the learning framework.
We program basic cooperative behaviors, such as agent gathering and joint attacking,
as cooperative actions in ECG.
Moreover, we unify both the primitive actions and the cooperative actions into the ECG graph with ECG target node, 
which fosters the seamless integration of both multi-agent reinforcement learning (MARL) and expert knowledge.

\item Thirdly,
ECG is a graph structure that naturally has stronger interpretability.
It is easy to visualize and monitor the cooperative behavior of agents
simply by observing the topology of ECG.
\end{enumerate}

% Pre-programmed with basic cooperative behaviors,
% the cooperative actions are the key component to blending cooperative knowledge into the learning framework.
Finally,
HCGL employs four graph operators, a group of virtual agents, to dynamically manipulate the ECG edge connections and change the topology of ECG in real time.
Since the environment is dynamically changing when agents carry out their cooperative operations,
it is necessary to adjust the topology of ECG in response to the environment.
To avoid ambiguity with real agents, 
we will refer to these virtual ECG agents only as graph \textit{operators} in the following sections.
Each graph operator possesses a policy neural network that is trained by an MAPPO \cite{yu2021surprising} RL learner,
and the sole goal of these graph operators is to maximize the reward of the entire multi-agent team.
These graph operators drive the cooperation graph 
to make adjustments and adaptations according to the changing environments,
thus guiding agents to accomplish their objectives.

The main contributions of this study are fourfold:
(1) We propose a novel hierarchical graph-based HCGL model with knowledge incorporation capability.
(2) We present an approach to visually interpret the behavior of agents with the topology of ECG.
(3) We demonstrate the effectiveness of ECG in large-scale cooperative tasks.
(4) We show that HCGL can efficiently transfer policies learned in small-scale tasks to large-scale ones.

In the following sections,
we first introduce related works on prior hierarchical single-agent RL and multi-agent RL.
Then we introduce our ECG framework beginning with the structure of a static cooperation graph.
Next, we emphasize how graph operators manipulate the graph to make it dynamic and responsive.
Finally, we perform solid experiments to justify the effectiveness of HCGL in different large-scale multi-agent environments.

%% file: tex/03-related.tex
\section{Related Works}

Research on multi-agent systems has been greatly inspired by the success of reinforcement learning. 
Unlike single-agent problems such as Go \cite{silver2016mastering} and RTS game \cite{vinyals2017starcraft, vinyals2019grandmaster}, 
multi-agent reinforcement learning (MARL) focuses on the development of cooperative policies among multiple agents.
The coordination among multiple agents poses a critical challenge in multi-agent learning
where agents engage in dynamic interactions to achieve a shared objective. 
E.g., Dota 2 \cite{berner2019dota}, SMAC \cite{usunier2016episodic}, multi-robot encirclement \cite{zhang2020multi}, 
and attacker-defender \cite{wu2021multi} scenarios. 

% limitation -> hierarchical RL
Recent studies suggest that achieving multi-agent coordination 
becomes progressively difficult as the number of agents increases \cite{samvelyan2019starcraft}.
% Recent studies suggest that the difficulty of achieving multi-agent coordination 
% increases exponentially with the increase in the number of agents \cite{samvelyan2019starcraft}.
While algorithms such as
Qmix \cite{rashid2018qmix},
QTRAN \cite{son2019qtran},
COMA \cite{foerster2018counterfactual},
ROMA \cite{wang2020roma} and
ACE \cite{li2022ace} achieve significant success in small and medium scale multi-agent problems,
it is discovered that existing MARL methods have many limitations in large-scale cooperative tasks.
For example, 
as the team size expands, it becomes challenging to accurately distribute rewards among individual agents. 
Moreover, the increase in the joint policy space exacerbates the difficulty of exploring agent actions.
Therefore,
it is imperative to investigate alternative approaches to tackle large-scale coordination problems. Examples of such attempts include hierarchical multi-agent reinforcement learning (Hierarchical-MARL), multi-agent curriculum learning \cite{wang2021survey}, and knowledge incorporation.

% hierarchical rl works
Hierarchical multi-agent reinforcement learning has gained popularity as a research area due to its capacity to address complex problems by partitioning them into sub-problems.
These methods usually improve the learning efficiency by re-engineering the agents' action space.
A study by \cite{spatharis2021hierarchical} proposes a 
hierarchical multi-agent reinforcement learning scheme for air traffic management, 
which divides the task into two levels: tactical and strategic.
Similarly, 
this work is extended by \cite{singh2020hierarchical} to maritime traffic management which models ships as individual agents and the VTS (vessel traffic service) as regulatory agents.
Other works extend this technique to air combat \cite{kong2022hierarchical},
UAV swarm confrontation \cite{wang2021uav, yue2022unmanned} and football simulations \cite{yang2019hierarchical}.
However,
most of these methods are task-specific and 
lack further generalization capability.

% multi-agent curriculum learning
Multi-agent curriculum learning aims to address difficult multi-agent problems
by learning easier tasks and increasing the difficulty according to a schedule.
For example,
start off by learning in a multi-agent setting with a limited number of agents, 
and then gradually and automatically increase the number of agents \cite{zhang2022automatic, wang2020few}.
This technique provides a solid solution to solve sparse-reward cooperative multi-agent problems \cite{chen2021variational}.
The potential applications of curriculum learning include
autonomous driving \cite{yang2021diverse}, SMAC \cite{wang2021uav, yue2022unmanned}, etc.

% Knowledge-driven
Another alternative approach is incorporating prior knowledge into the learning algorithm.
\cite{deng2020integrating} discussed a variety of ways to use general human knowledge to facilitate
reinforcement learning.
\cite{singla2019memory} proposes a memory-based RL model that associates the learning process
with environmental knowledge to solve UAV obstacle avoidance problems during navigation.
\cite{han2020improving} suggests a fresh strategy for enhancing the efficiency of multi-agent reinforcement learning 
by employing logic rules, a prevalent form of human knowledge, to address SMAC challenges.
A more general form of the knowledge-driven approach is knowledge graph reasoning,
which is used in other RL problems such as recommendation \cite{xian2019reinforcement, wang2022multi}.

%% file: tex/04-method.tex
\section{Extensible Cooperation Graph}
\begin{figure}[!t] 
    \centering
    \includegraphics[width=\linewidth]{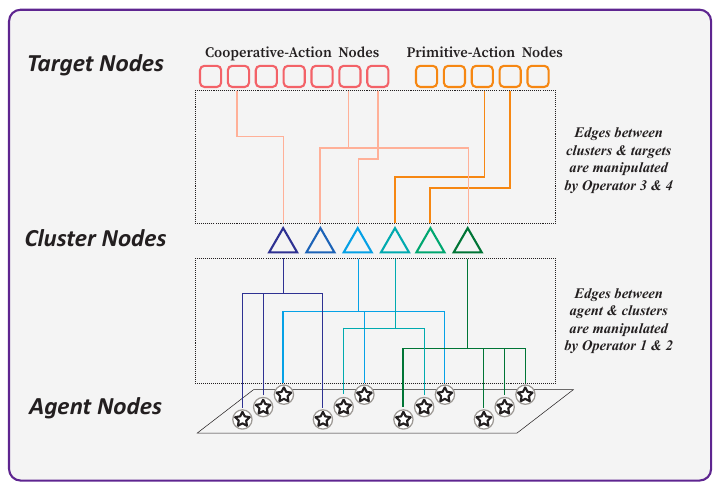}
    \caption{
        An illustration of the Extensible Cooperation Graph (ECG).
        Starting from the bottom, 
        ECG is a three-layer hierarchical graph structure that includes agent nodes,
        cluster nodes and target nodes.
        The graph nodes are connected by edges.
        Within each episode,
        this graph will be dynamically controlled by four virtual agents referred to as operators.
    }
    \label{fig:ECG_main}
\end{figure}

We first introduce the core structure of HCGL: the Extensible Cooperation Graph (ECG).
ECG is a multi-layer non-neural graph with a simple topology (directed and acyclic).
It acts as a mastermind to control the collective behavior of agents 
by changing its own edge connections dynamically.
A basic ECG $G$ contains three layers: the agent layer ($\mathcal{A}$), 
the cluster layer ($\mathcal{C}$) and the target layer ($\mathcal{T}$), respectively.
% $G=(V,E)$

% Agent 的特殊点
An agent node $a_i \in \mathcal{A}$ represents an agent in the environment.
Unlike any other MARL algorithm,
each agent in the graph acts as a pawn and does not possess a policy neural network of its own.
Instead,
the behavior of agents depends on the dynamic graph structure and edge connections 
(which will be discussed in the next section).

% Cluster 介绍
The cluster nodes are responsible for dividing agents into groups,
which organize agents to complete difficult multi-agent task objectives.
A cluster node $c_k \in \mathcal{C}$ can contain all agents when needed.
On the other hand, each agent is required to join only one cluster at a specific time step.
When an agent $a_i$ joins a cluster $c_k$,
we use a directed edge in ECG to represent this relationship $(a_i, c_k)$.

% Target 介绍
% The most essential target nodes  cooperative cluster actions 
The target nodes represent special actions that need to be executed by clusters.
The graph relationship of clusters and targets is very similar to that of agents and clusters,
and each cluster chooses one of the targets to follow at each time step.
When a cluster $c_k$ selects a target $t_\tau \in \mathcal{T}$,
an edge is established, directing from itself towards the target $(c_k, t_\tau)$.
There are two types of target nodes, namely the primitive-action nodes and cooperative-action nodes.

We introduce primitive-action nodes as the first type of target nodes.
Each action from the original agent action space $\mathcal{U}$ is represented as a target node in ECG.
When a cluster is connected to a primitive-action target node,
all of its member agents perform corresponding the primitive actions simultaneously.
In other words,
the primitive actions are broadcasted from its target node to all connected clusters,
and eventually to all agents connected.

Next, we introduce cooperative-action nodes, the second type of target nodes.
Unlike primitive actions from the original agent action space $\mathcal{U}$,
cooperative actions are executed by clusters instead of agents.
Consequently, 
a cooperative action can simultaneously affect all agents in a cluster.
When a cluster is connected to a cooperative-action  node,
the cluster needs to parse the cooperative action first according to the status of its member agents,
translating cooperative actions into primitive actions for each individual agent  
before finally distributing the resulting primitive actions to each member agent.
As an example, 
the ``gathering'' and ``scattering'' are two simple cooperative actions.
When a cluster is performing the ``gathering'' action,
this cooperative action is translated to primitive actions that move each agent towards its allies.
And ``scattering'' cooperative action plays the opposite role and moves agents away from one another.
The same cooperative action is usually interpreted into different primitive actions 
for different agents within a cluster.

% Both raw-actions and coop-actions are important in ECG.
% raw actions define basic agent behaviors while 
% coop-actions extend the choice of actions.

Cooperative actions are essential for integrating cooperative knowledge into the MARL framework.
Without the participation of cooperative-action nodes,
learning joint actions like gathering can be time-consuming 
if agents are not guided by dense rewards from the environment.
And even if agents finally learn to perform similar skills, 
the policies still lack transparency and interpretability.
The introduction of cooperative-action nodes provides a framework where
knowledge-based joint behaviors 
have the opportunity to take part in the policy optimization process.
Moreover,
since the edge connection of ECG is dynamic and 
the adjustment policies are gradually optimized during the learning session,
the framework tolerates not only the co-existence of primitive actions and cooperative actions,
but also the co-existence of different cooperative actions with conflictions.
Detrimental cooperative actions can be easily eliminated during learning
by treating them as unwelcomed nodes.

% link with pure rule
Furthermore,
the ECG framework provides a compromise between pure RL methods 
and pure pre-programming methods (rule-based).
Many abstract cooperative behaviors, such as ``joint attacking'' 
and ``rallying at a certain location'' from rule-based methods, can be utilized as cooperative actions in ECG.
At the same time,
these cooperative actions share an equal position with the primitive actions in ECG.
This means that if a cooperative action is not helpful in addressing the problem in the current environment,
the corresponding node will be gradually given up during reinforcement learning.

\section{Hierarchical Knowledge Embedding}
% Cluster Coop Action as a media to hold collabrative knowledge embeddings
In our HCGL model,
target nodes are the main media to integrate knowledge with MARL.
There are two types of target nodes that represent the cooperative actions and the primitive actions, respectively.

% - primitive and 
The primitive action corresponds to actions in classic MARL algorithms,
where agents need to choose one of the primitive actions to interact with the environment.
Cooperative action, on the other hand, is a new concept proposed in HCGL, and it differs from primitive actions in multiple ways.
First and most importantly, 
these actions are executed by clusters rather than agents.
A cooperative action guides the collective behavior of all agents in the cluster executing it, e.g., making a group of agents maneuver and generate a formation.
Secondly,
each cooperative action is programmed with expert knowledge,
allowing us to integrate low-level cooperative skills into the reinforcement learning framework.
Thirdly,
the cluster nodes will eventually translate each cooperation action into different primitive actions for each agent individually,
considering the purpose of the cooperation action and the state of agents in the cluster.

From the perspective of ECG,
when a cluster node is connected to a primitive action,
the primitive action will be passed to all agents linked to this cluster unchanged.
In comparison,
when a cluster node is connected to a cooperative action,
the cooperative action is first translated to primitive actions before
being delivered to each agent.

The introduction of cooperative actions enables a variety of ways to incorporate expert knowledge:
\begin{itemize}
    \item Employing basic cooperation knowledge such as gathering, scattering, encirclement, following, patrolling, etc.
          These cooperative behaviors are not only simple but also general in a variety of tasks.
          Unfortunately, learning these simple behavior may take millions of steps in non-hierarchical RL models.
          In comparison, it only takes a few lines of code to write a robust controller to achieve these basic behaviors.
          In the HCGL, we program these basic behaviors as cooperative actions to accelerate the learning process.
    \item Applying additional constraints such as flight formation, collision avoidance, keeping communication range, etc.
          Agents trained in simulations often deviate from our expectations because many factors are not considered in the simulation environments. For example, although we have the knowledge that a V-shaped flight formation can reduce flight energy costs. But such behavior cannot be learned unless we consider aerodynamic factors in the simulation.
          Likewise, many simulation environments do not detect collisions to reduce computational costs,
          and meanwhile, there are many existing collision avoidance controllers that can be leveraged.
          In such cases,
          our HCGL model can take full advantage of these existing algorithms and make them an integrated part of the multi-agent cooperation policy.
\end{itemize}

Note that the HCGL model not only passively incorporates knowledge-based cooperative actions but also actively evaluates, chooses, and switches between them.
If some cooperative actions are useless, detrimental, or in conflict with one another,
the effectiveness of HCGL is not significantly impacted
because the topology of ECG is adjusted by graph operators.
The graph operators can distinguish detrimental actions and disconnect clusters from these action nodes during the MARL training stage.
The following section introduces the method for adjusting ECG.

% If a problematic cooperative action influences the performance of the team,
% the graph operators can learn to

% each cooperation action is eventually translated to primitive actions
% considering the state of each agent individual in the cluster.
% A cluster node that receives a cooperative action will
% eventually determine and deliver primitive actions back to agent nodes though ECG.

% each cooperative action uses primitive actions as its foundation,
% A cluster receiving a gathering cooperative action will
% eventually deliver the movement-related primitive actions back to agents though ECG.

% we can use following actions

\section{Learning Graph Manipulation}

% 引出四个operator
As we have mentioned in the previous sections,
the agents (agent nodes) do not possess policy neural networks,
they only receive instructs from the ECG which must change itself dynamically in each episode.
The adjustment of ECG will be carried out by a group of virtual agents referred to as Graph Operators.
To distinguish them from other regular agents manipulated inside ECG,
we will refer to them only as \textbf{operators} for clarity.
There are always four operators to jointly manipulate a ECG.
Two of them ($o_{p1}$, $o_{p2}$) are responsible for adjusting the edges between agents and clusters,
while the other two ($o_{p3}$, $o_{p4}$) are responsible for adjusting the edges between clusters and targets.

% 动作空间
The graph operators have a special action space since their job is
to manipulate the ECG rather than directly interact with the environment.
To transfer an agent from one cluster $c_m$ to another cluster $c_n$,
operator $o_{p1}$ and operator $o_{p2}$ work together to produce $m \in |\mathcal{C}|$ and $n \in |\mathcal{C}|$ respectively.
There are two special cases.
First, if operator $o_{p1}$ chooses a cluster that is currently empty (with no agent nodes linking to it),
the action of both $o_{p1}$ and $o_{p2}$ is invalid, regardless of the action of $o_{p2}$.
Second, if operator $o_{p1}$ and $o_{p2}$ happen to choose the same cluster,
their action is invalid as well.

Similarly,
the action space of the other two operators $o_{p3}$ and $o_{p4}$
controll the edges between clusters and targets in the same way.
Operators $o_{p3}$ and $o_{p4}$ produce a pair of actions to transfer
a cluster linked with target node $t_m$ to a different target node $t_n$,
where $m \in |\mathcal{T}|$ and $n \in |\mathcal{T}|$.
Again, if no cluster is currently connected to $t_m$ or if $t_m$ and $t_n$ are the same target,
this action pair is invalid.

\begin{figure*}[!t] 
    \centering
    \includegraphics[width=\linewidth]{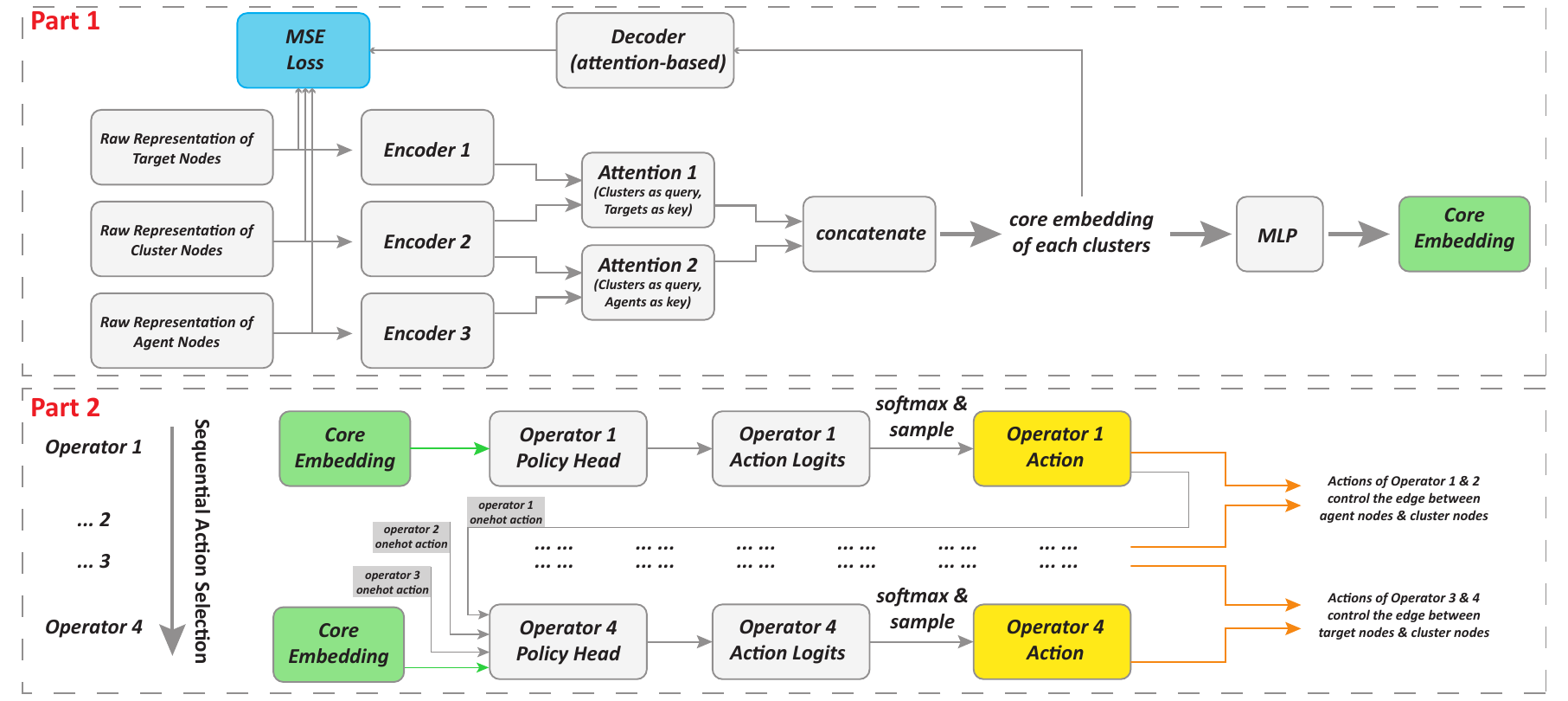}
    \caption{ Training the policy network of graph operators. }
    \label{fig:PolicyNetwork}
\end{figure*}

% 动作屏蔽
As the number of target nodes is significantly 
greater than the number of clusters,
the operators are likely to spend much time exploring action combinations 
that are ineffective.
To address this issue,
we perform action masking to forbid operators from choosing invalid actions 
by providing operators with their current available actions.
More specifically,
this regulation is imposed on $p_{1}$ and $p_{3}$.
 
\section{Training Graph Operators}
\subsection{Training Operators}
% 参数与算法
Graph operators are trained with multi-agent RL methods MAPPO, 
which treat operators (instead of agent nodes) as \textbf{agents}.
Like many other multi-agent systems,
each graph operator has its own policy network with shared parameters.
As illustrated in Fig.\ref{fig:PolicyNetwork}, 
most policy network parameters are shared among graph operators.
Nevertheless,
each operator possesses a unique policy head to produce different graph manipulation actions.

% 决策依赖
We previously introduced that operator actions are only valid when considered in pairs,
so it is important that an operator is aware of the action of its fellow operators before making the final action decision.
Inspired by ACE \cite{li2022ace}, 
in our model, the operators sequentially select their actions at each step.
Operator 1 (op.1) is the first operator to select an action.
When op.2 determines its own action, it refers to the action of op.1.
Similarly, op.3 and op.4 take all previous operator actions into consideration when selecting their own actions.

\subsection{Initialization of ECG}
% random at beginning, freeze afterward
Operators must learn to manipulate ECG differently to generate distinct ECG topologies to perform different cooperative actions.
However, the initial ECG topology in each episode still needs to be determined.
We adopt the method introduced in \cite{fu2022cooperation}, 
where the ECG initial topology is generated randomly at the beginning of training.
Then, this topology is frozen as the initial state of ECG for all future episodes.

% 用高熵拓扑
As suggested by \cite{fu2022cooperation},
although the initialization of ECG topology is random,
we can still improve the stability of the model by producing multiple random topologies,
and then choosing one of them according to the entropy of the edge connections.
% $$
% H_{\text{ECG}} = 
% $$,
% where $H_{\text{ECG}}$ is defined as the ECG entropy, ....
This is because there may exist multiple sub-optimal topologies across all possible ECG structures. 
We can measure the distance between topologies in terms of 
the number of operator steps required to complete the transition between two topologies.
A high-entropy initial topology can increase its distance from sub-optimal topologies,
encouraging graph operators to conduct more exploration before converging.
A more comprehensive exploration can aid graph operators in identifying optimal policies more efficiently.

\subsection{Operator Policy Stability Enhancement.}
Operators have significant privileges in our model, 
as they can alter the behavior of the entire team with minimal effort.
This feature of ECG
offers considerable hierarchical advantages concerning the training of agents.
However, it may potentially weaken the connection between the environment and graph operators.

% autoencoder
We propose two improvements to enhance the bond between graph operators and the external environment.
On one hand,
we introduce an autoencoder to connect the operator observation input and 
the representation produced by the last parameter-sharing layer.
A typical autoencoder has an encoder that maps input to a low-dimensional space
and a decoder that reverses the encoding process 
to estimate the original input from the low-dimensional representation.
In this model,
the parameter-sharing feed-forward policy layers perform the role of the encoder,
and we only need to add an additional decoder to map the final representation
to estimate the original operator observations.
The loss function (mean squared error) $\mathcal{L}_{ae}$ (together with policy gradient loss) 
is applied during the policy network update.

% inferences
On the other hand, 
we generate interference on ECG by altering the ECG edge connections against the action of operators.
Similar to the dropout technique utilized in other machine learning problems, 
interference is applied randomly at the start of every step with a small probability denoted by $p_i$.
We select $p_i=0.5\%$ by default.
More specifically,
a set of random but fake operator actions is produced and applied to the ECG when interference is triggered.
With the participation of interference,
operators are forced to learn to reverse the influence caused by fake operator actions.
Moreover,
this technique motivates operators to learn to analyze the representation of ECG,
facilitating the ability to understand the relationship between the environment and ECG itself.

\begin{figure*}[!t] 
    \centering
    \includegraphics[width=0.9\linewidth]{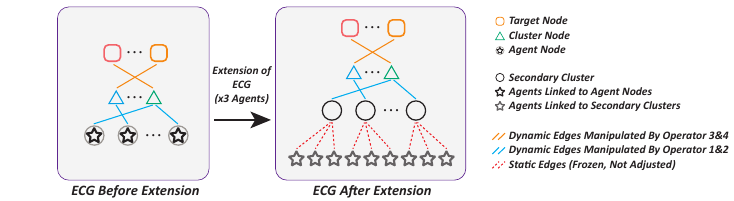}
    \caption{
        ECG has excellent scaling capability and has significant advantages 
        when used together with curriculum learning techniques.
    }
    \label{fig:GraphExtension}
\end{figure*}

\section{Policy Network of Graph Operators}

In our ECG model, 
the graph operator agents (instead of environment agents) are trained by reinforcement learning.
As a result,
the training process differs from a typical actor-critic RL model.

\subsection{Observation.}
Graph operators gather available information on all nodes in ECG, 
including agent nodes, cluster nodes, target nodes and ECG's edge connection.
As illustrated in Fig.\ref{fig:PolicyNetwork},
the raw representations of agent nodes are simply the observation of each agent.
On the other hand,
The raw representations of clusters are one-hot vectors.
And the raw representations of each target node include available information about this target.
If the target node is a primitive action, its representation is one hot,
and if the target node is a coop-action, 
information about this coop-action is placed in the representation instead of one-hot coding.
For instance, if the coop-action concerns a rally point, 
the location of this point is placed into the raw representation.

\subsection{Forward Network.}
% We adopt attention module as the building block for operator policies and autoencoders.
As shown in Fig.\ref{fig:PolicyNetwork},
the representations of targets, clusters and agents are normalized
and then projected into embeddings with fixed dimensions, respectively.
Next,
we compute attention between cluster nodes and agent nodes (Attention-AC)
and cluster nodes and target nodes (Attention-CT).
In this step,
we take the representation of clusters, the most important nodes in ECG, as attention queries,
and the representations of agent nodes as well as target nodes as attention keys.
More specifically,
both attention layers are cluster-oriented.
Attention-AC captures the graph relationship between agents and clusters,
combining the representation of each agent with the representation of clusters.
On the other hand,
Attention-CT captures the graph relationship between targets and clusters,
gathering the feature from target nodes to cluster nodes.
Then the attention results from Attention-AC and Attention-CT
are concatenated and further processed by a Fully Connected (FC) layer,
which extracts an embedding $e_h$ that holds a comprehensive ECG representation.
The resulting $e_h$ values are eventually utilized for three tasks: 
(1) estimating the state value, (2) estimating the action logits, 
and (3) reconstructing the raw node representation 
using the decoder component of the autoencoder.

For purposes (1) and (2),
the embedding of all clusters will be concatenated together,
then produce a low-dimension embedding via FC layers.
The critic uses this embedding to produce the state value estimation.
Each graph operator also uses this embedding sequentially to determine the operator action (also via FC layers).

For purpose (3),
we reconstruct raw node representations with additional attention layers:
Attention-AE-A, Attention-AE-C and Attention-AE-T.
Using one-hot vectors as query and $e_h$ as keys,
the reconstructed agent node representations are computed by Attention-AE-A,
then obtain the reconstruction loss with MSE.
The reconstruction loss of clusters (reconstructing with Attention-AE-C) and 
targets (using Attention-AE-T) is produced likewise.

% \subsection{Projecting to Operators' Action Space}

\section{System Scaling by Curriculum Learning}

% duplicating the agent nodes
This section proposes a curriculum learning method that leverages the extensibility of ECG to handle large-scale multi-agent problems. 
The method starts with simple environments featuring fewer agents 
and gradually transfers to complex environments requiring a large number of agents.

% To achieve the expansion of team,
ECG supports the expansion of the graph by converting its agent nodes to secondary cluster nodes.
Unlike its predecessor agent node that only controls a single agent, 
a secondary cluster node connects 2 or even more agents,
thus expanding the size of the team several times larger.
After the expansion,
the graph between targets and clusters and the policy of op.3 and op.4 stay unchanged.
And op.1 and op.2 are now responsible for controlling the edge between clusters and secondary clusters,
using the same policy they learned in the smaller-scale environment.
Finally, the connection between secondary clusters and the expanded agent team is considered static in ECG,
this part of the graph will not be adjusted like the rest of ECG.

% policy observation
The change in ECG also impacts the observation of the policy network of operators.
The secondary cluster now connects to the observation of multiple agents rather than a single one.
However,
the policy network of graph operators still accepts same-shape observation from only one node.
In order to successfully transfer the policy parameters learned in the previous task to the larger scale new task,
we add an extra attention layer at the beginning of the policy network 
to merge the observation embeddings of all agents connected to a secondary cluster into one.
The initial parameters of the new attention layer are randomly initialized.
In contrast,
the parameters of the rest of the network inherit from the previous policy network learned in the previous environment.

% keep learning
% Like most curriculum learning problems,
% it is likely that the policy of operators learned in previous environment
% is not fully capable of commanding agents in a larger team.
% Moreover, 
% the new attettion layer added for compatibility also requires training.
% Thus,
% operators need to learn new adaptive policies based on old policies learned in previous scenerios.

Like most curriculum learning scenarios, 
it is likely that the policy of operators learned in the previous environment 
is not fully capable of commanding agents in a larger team. 
Moreover, the new attention layer added for compatibility also requires further training. 
Thus, building on the ones learned in previous scenarios,
operators need to learn new adaptive policies and further optimize their policies with RL.

%% file: tex/05-experiment.tex
\section{Simulations}

\begin{figure}[!t] 
    \centering
    \includegraphics[width=\linewidth,height=0.6\linewidth]{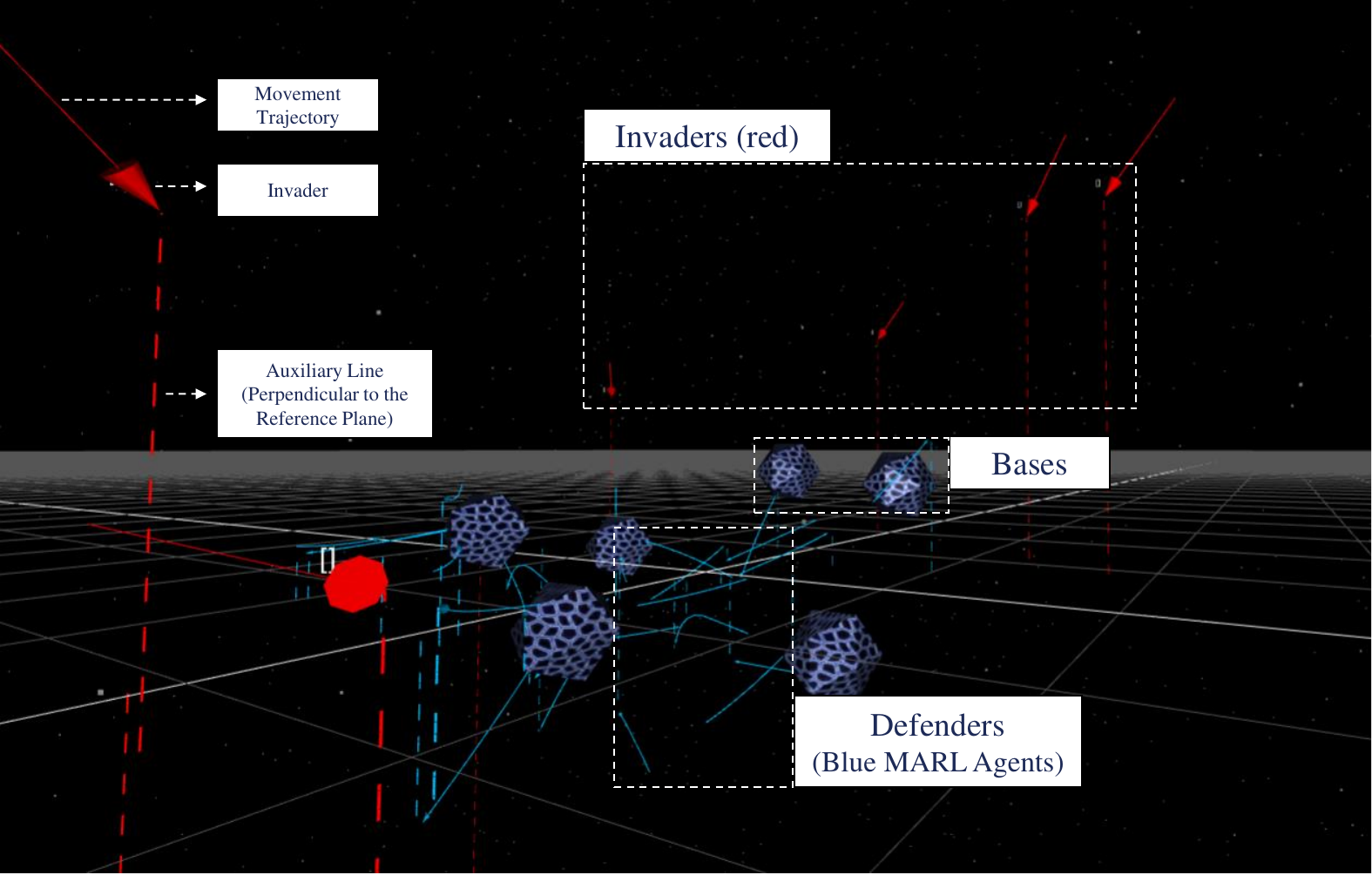}
    \caption{ 
        Cooperative Swarm Interception (Initial State).
    }
    \label{CSI}
\end{figure}
\begin{figure}[!t] 
    \centering
    \includegraphics[width=\linewidth,height=0.6\linewidth]{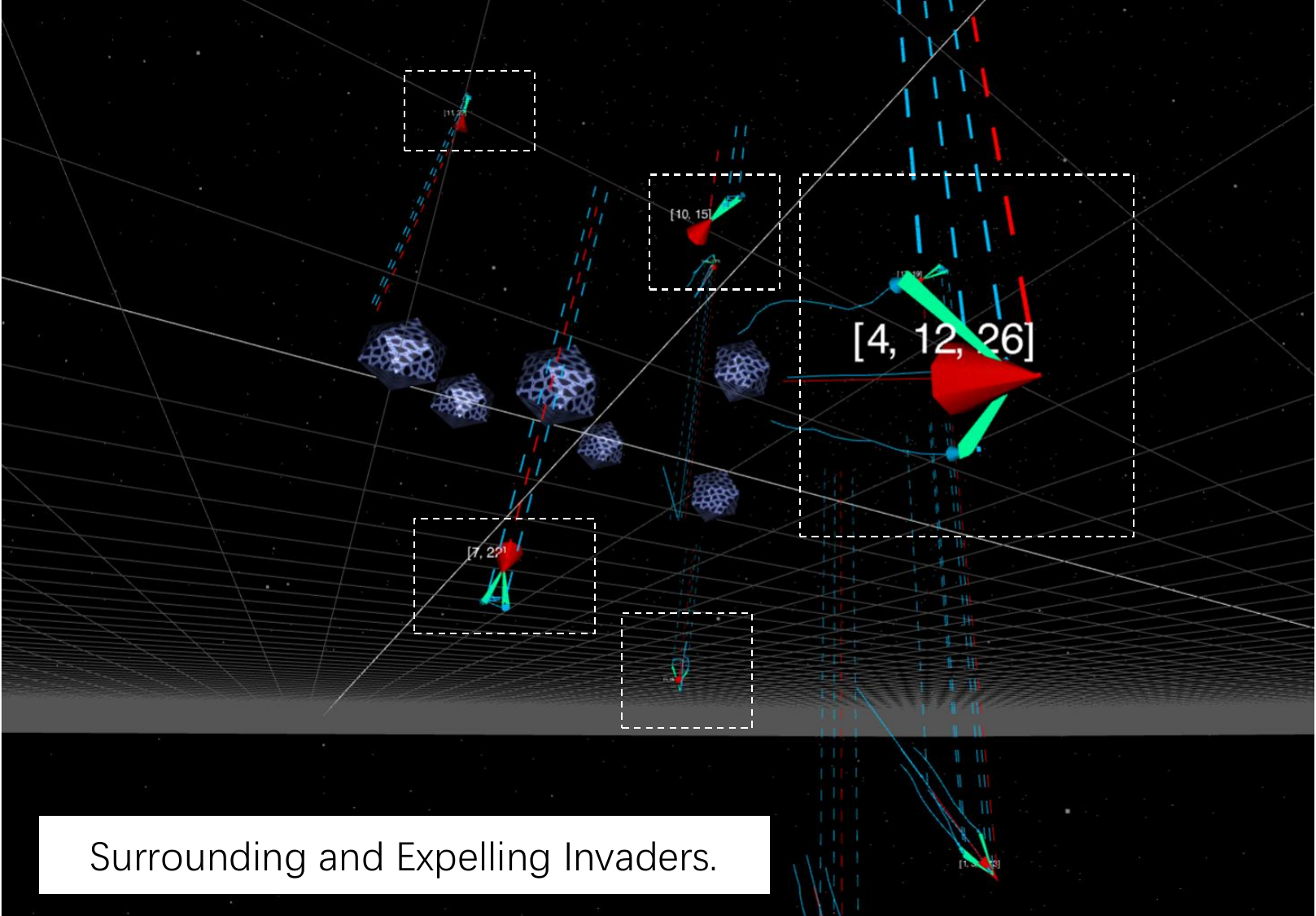}
    \caption{ Cooperative Swarm Interception (Expelling invaders).
    }
    \label{CSI2}
\end{figure}

In this section, 
we evaluate the performance of HCGL in the Cooperative Swarm Interception benchmark.
We compare HCGL with existing MARL algorithms.
The algorithms evaluated are:
\begin{itemize}
    \item HCGL. The proposed method in this paper.
    It has two primary hyper-parameters,
    namely the number of clusters and the selection of cooperative actions.
    \item Qmix \cite{rashid2018qmix}. A value-based MARL algorithm 
    that leverages the Individual-Global-Max (IGM) principle
    and proposes a reward decomposition method.
    Qmix has SOTA performance in multiple MARL baselines.
    \item Qtran \cite{son2019qtran}. 
    Qtran improves Qmix by providing more general ways of decomposition.
    \item Qplex \cite{wang2020qplex}. 
    Qplex proposes a duplex dueling network architecture to factorize the joint value function.
\end{itemize}

\subsection{Cooperative Swarm Interception.}
% We use a sparse reward benchmark, Cooperative Swarm Interception (CSI),
% to evaluate the effectiveness of our ECG model.
% In this scenario,
% a team of defender agents need to prevent multiple invading hostiles from
% destorying their base.
% As shown in Fig.\ref{CSI},
% an invader can be shown down when it is surrounded by enough number of agents.
% Furthermore, when the number of agents tracking an invader exceeds a threshold,
% the invader will be force to change its direction away from its target,
% thereby eliminating the threat of the invader.
% For example, CSI-54/6/9 means 54 defender agents intercept 9 invaders,
% abd each invader can be force to turn back by at least 6 agents.

The effectiveness of our ECG model is evaluated using a sparse reward benchmark called Cooperative Swarm Interception (CSI). The scenario involves a team of defender agents that must prevent multiple hostile invasions from destroying their bases. 
As illustrated in Figure 1, an invader can be slowed down when it is surrounded by a sufficient number of agents. Moreover, if the number of agents tracking an invader surpasses a specific threshold, the invader is compelled to change its direction away from its target, thus neutralizing the threat of the invader.
For example, CSI-54/6/9 denotes that there are 54 defender agents to intercept nine invaders, and each invader can be forced to turn back by at least six agents.

The CSI benchmark uses extremely sparse rewards.
The agents are rewarded +1 as a team when all bases survive the attack.
If any base is destroyed by any invader,
the episode is considered failed and the agents are rewarded -1.

For HCGL, we program two types of additional cooperative actions:
intercepting one of the invaders and defending one of the bases.
The number of the cooperative-action nodes is the sum of the number of invaders and bases.

\subsection{Setup.}
The experiments are performed on regular x86-64 servers with GPUs.
MAPPO \cite{yu2021surprising} optimizer is used to train the policy of the graph operators.
The learning rate of MAPPO is $1\times 10^{-4}$,
the batch size is 128 (episodes),
the dimension of hidden embeddings is 64,
the entropy loss coefficient is 0.01,
and the epoch in each PPO optimization step is 16.

The implementation of Qmix, Qtran and Qplex is based on the PyMarl2 \cite{hu2021rethinking} platform,
which has fine-tuned algorithms such as Qmix and obtained SOTA performance in multiple benchmarks.
The batch size is also 128 episodes,
but note that the value-based methods have very different policy optimization cycles from MAPPO,
as their policy updates are much more frequent even when the batch size is identical.

The most important hyper-parameters in HCGL are the number of clusters $n_k$.
We use $n_k=14$ by default and provide an ablation study on the selection of $n_k$ in the following sections.

\subsection{Learning ECG Operator Policies.}

\begin{figure}[!t]
\begin{subfigure}{.49\linewidth}
\centering
\includegraphics[width=\linewidth]{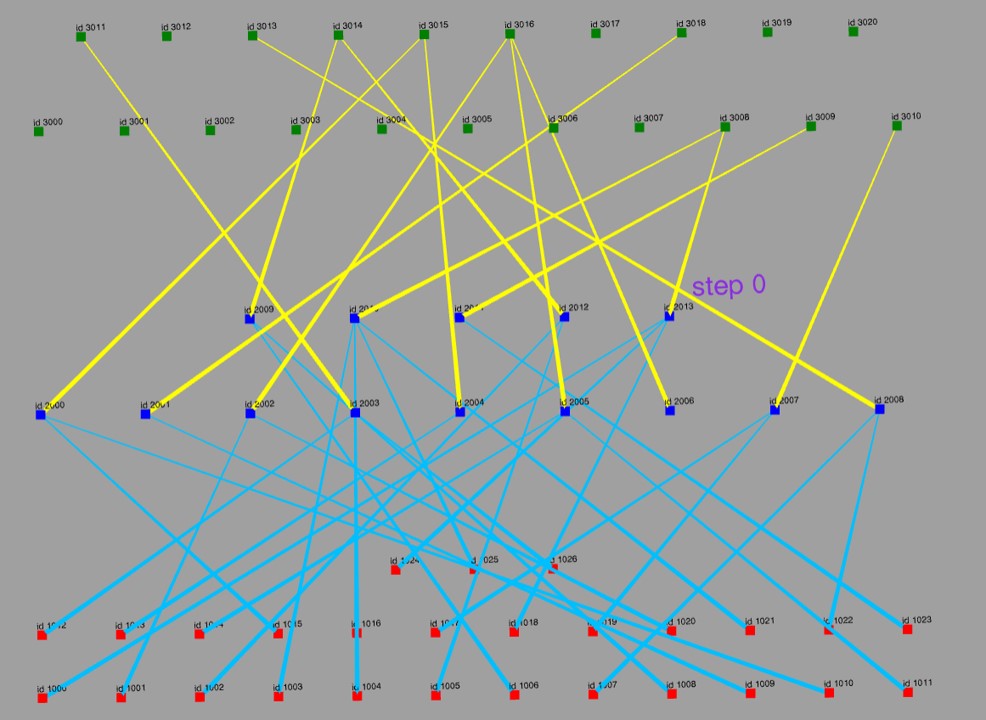} 
\caption{
ECG topology at step 0.
}
\label{fig:a}
\end{subfigure}
\begin{subfigure}{.49\linewidth}
\centering
\includegraphics[width=\linewidth]{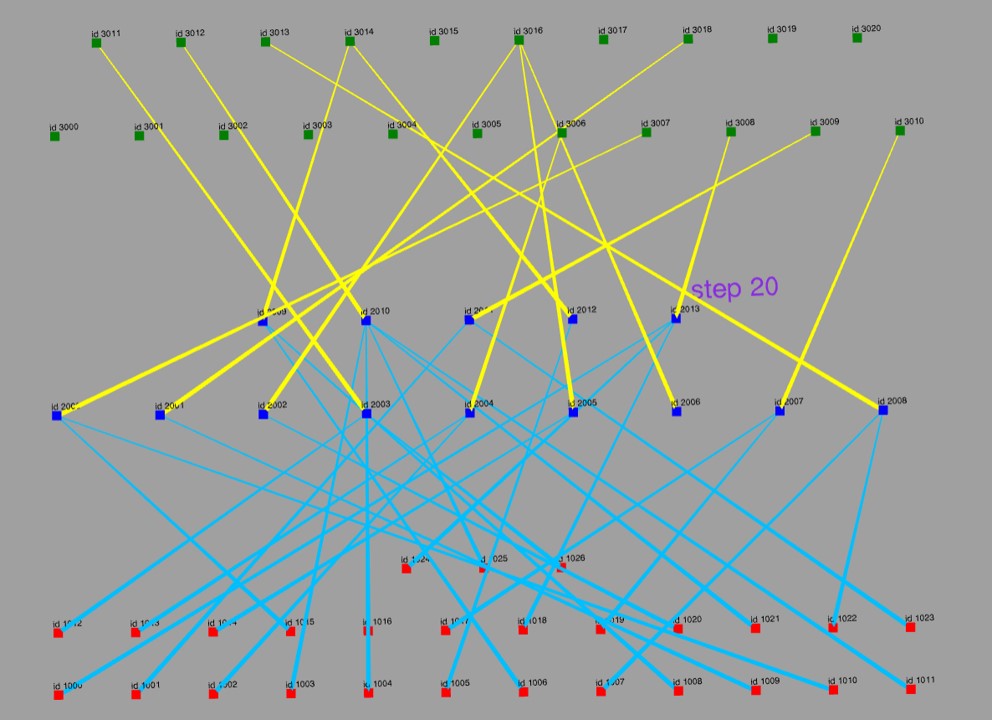} 
\caption{
ECG topology at step 20.
}
\label{fig:b}
\end{subfigure}

\begin{subfigure}{.49\linewidth}
\centering
\includegraphics[width=\linewidth]{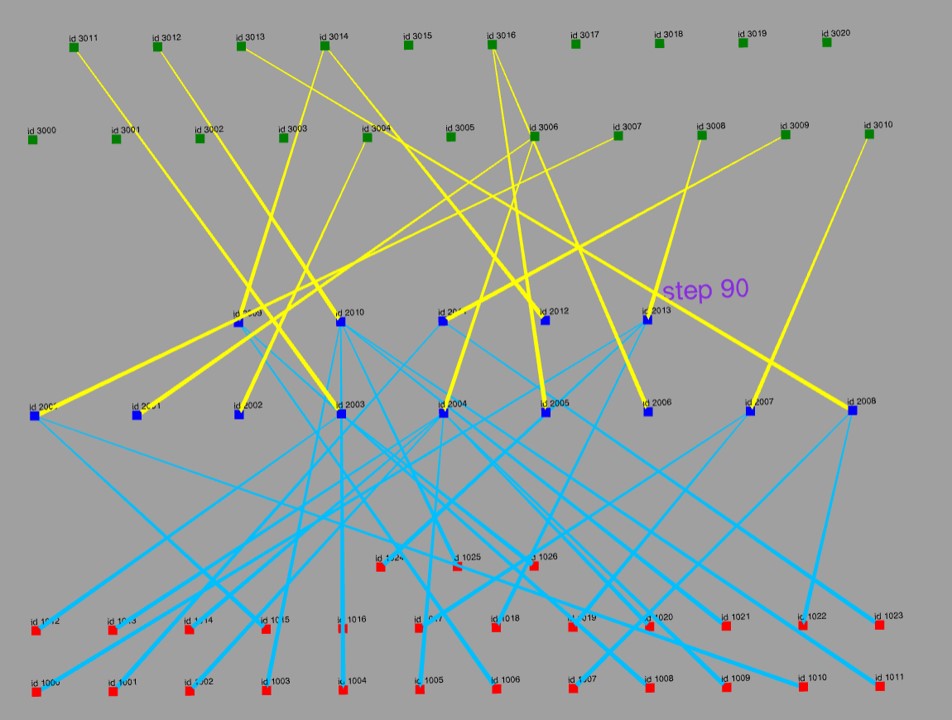} 
\caption{
ECG topology at step 90.
}
\label{fig:c}
\end{subfigure}
\begin{subfigure}{.49\linewidth}
\centering
\includegraphics[width=\linewidth]{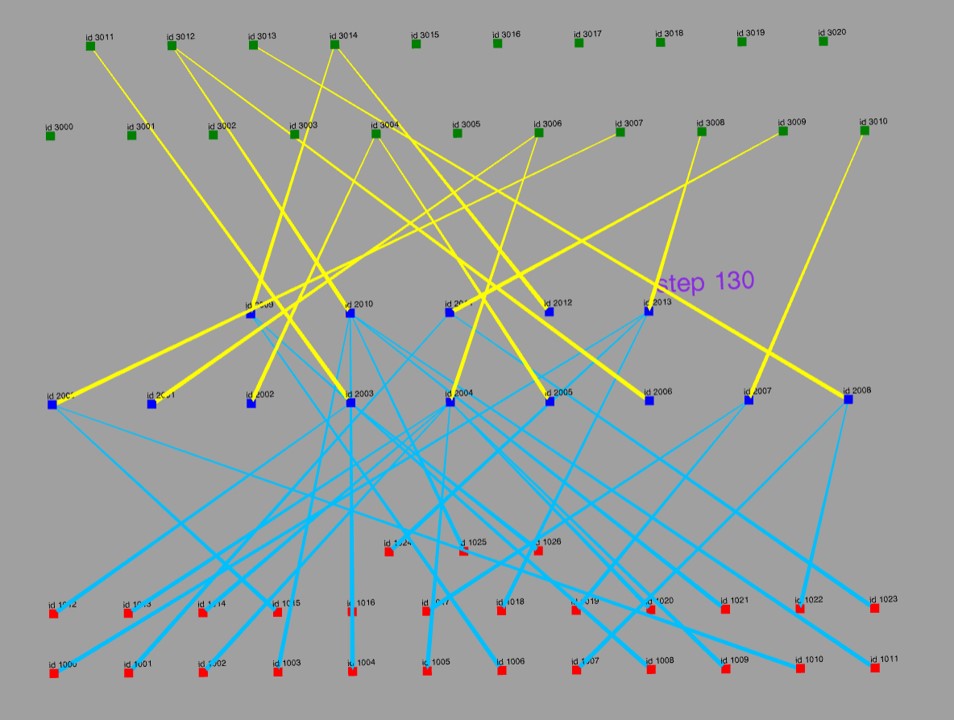} 
\caption{
ECG topology at step 130.
}
\label{fig:d}
\end{subfigure}

\caption{ECG topology in an episode in task CSI-27/3/9.
The positioning of the nodes in the figure is as follows: 
Agent nodes are depicted in red and positioned at the bottom, 
Cluster nodes are shown in blue and positioned in the middle, 
while Target nodes are illustrated in green and positioned at the top.
And the edge connections are represented by blue and yellow lines
in this figure.
}
\label{fig:topology}
\end{figure}

\begin{table}
    \caption{Comparing HCGL with other MARL baselines. 
    And testing the transferability of the learned HCGL policy.}
    \label{sample-table}
    \centering
    \begin{tabular}{ccc}
    \toprule
    {Method \& Task} &  {Zero-Shot Success Rate} & {Final Success Rate} \\
    \midrule
    Qtran CSI-27/3/9  &  -                                  &  0.00\tiny{$\pm$0.00}  \\
    Qplex CSI-27/3/9  &  -                                  &  0.00\tiny{$\pm$0.00}  \\
    Qmix CSI-27/3/9   &  -                                  &  0.00\tiny{$\pm$0.00}  \\
    \textbf{HCGL CSI-27/3/9}    &  -                & \textbf{0.97}\tiny{$\pm$0.03}  \\
    HCGL-TF CSI-54/6/9    &  \textbf{0.77}\tiny{$\pm$0.06}       &  0.87\tiny{$\pm$0.03}  \\
    HCGL-TF CSI-81/9/9    &  0.75\tiny{$\pm$0.03}                &  0.83\tiny{$\pm$0.04}  \\
    HCGL-TF CSI-108/12/9  &  0.71\tiny{$\pm$0.06}                &  0.84\tiny{$\pm$0.04}  \\
    HCGL-TF CSI-216/24/9  &  0.65\tiny{$\pm$0.09}                &  0.95\tiny{$\pm$0.04}  \\
    \bottomrule
    \end{tabular}
    \end{table}

This experiment evaluates the performance of HCGL with CSI benchmark,
and gives a comparison between HCGL and other SOTA methods.
The success rate metric is used for assessing the performance of algorithms.

The results are displayed in Table \ref{sample-table}.
Although we have tried multiple hyper-parameter settings,
Qtran, Qplex and Qmix methods are unable to solve the CSI benchmark problem at all.
In contrast,
our HCGL can achieve a {0.97}{$\pm$0.03} success rate,
significantly outperforming previous methods.

In order to demonstrate the ECG topology shift during the task,
we record the edge connections in one of the episodes and display the ECG topology in Fig.\ref{fig:topology}.
The agent nodes, cluster nodes and target nodes are placed from bottom to top
with red, blue and green respectively.
The connection of ECG is dynamically changing according to the game situation.
% Most of the clusters are connected by agents,
% but the number of connected clusters reduces after 90 step.
% The edges between clusters and agents are changing simutanously with the edges between clusters and targets.
The majority of clusters are linked by agents; nevertheless, the number of connected clusters dwindles after 90 steps. 
These un-connected clusters suggest that some of the clusters are redundant. 
It is also suggested that the connectivity of edges between clusters and agents changes simultaneously with the connectivity of edges between clusters and targets.

% 水字数
% At each step, each agent observes
% reward
% action space
% set performance metrics

% outperform
% The reason is that
%  Firstly,  outperforms which shows the 

\subsection{Transferability.}

\begin{figure}[!t] 
    \centering
    \includegraphics[width=\linewidth]{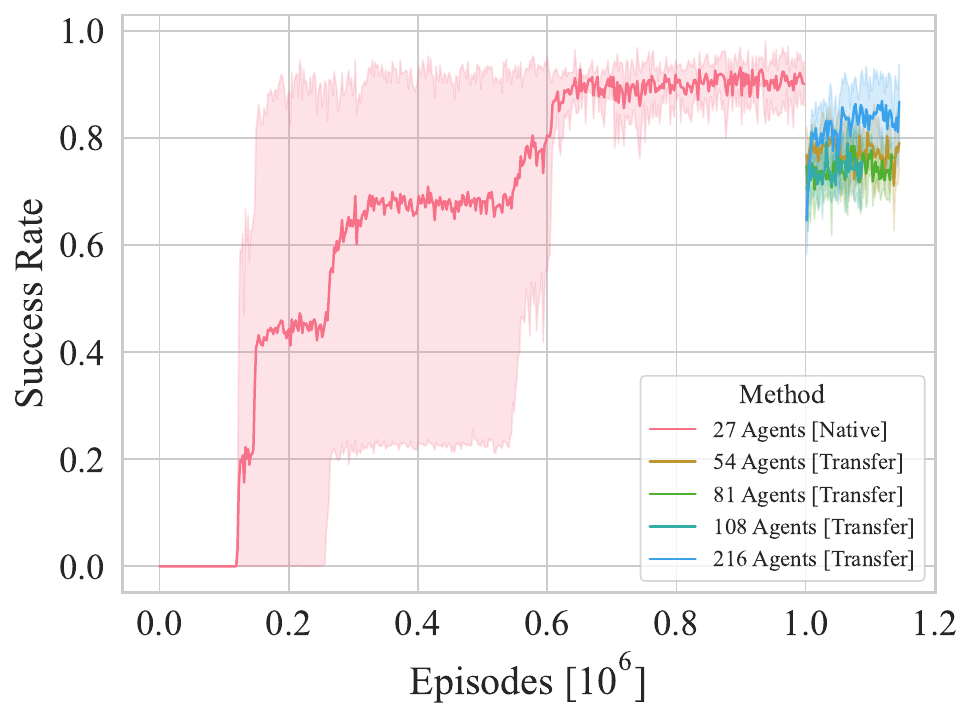}
    \caption{
        Testing transferability of HCGL. 
        The policy is firstly trained in task CSI-27/3/9,
        then the trained policy is transferred to CSI-54/6/9, CSI-108/12/9, etc.
    }
    \label{fig:Transferability}
\end{figure}

Benefits from the hierarchical ECG structure,
HCGL has significant transferability for large-scale multi-agent environments participated by hundreds of agents.
One of the limitations of non-hierarchical multi-agent reinforcement methods is their poor performance in transferability.
Most existing MARL algorithms have to be retrained
when the number of agents changes because 
their number of policy parameters is related to the number of agents.
Previous studies have attempted to improve compatibility by leveraging the Transformer structure or attentional graph network.
However, a non-hierarchical structure is insufficient for transferability 
when the number of agents increases twofold or threefold.

In this simulation,
we first train the policy in CSI-27/3/9 from scratch,
and transfer the trained policy to CSI-54/6/9, CSI-81/9/9, CSI-108/12/9,
and CSI-216/24/9.
The training process is depicted in Fig.\ref{fig:Transferability},
and the result is displayed in Table \ref{sample-table}.
The transferred methods are denoted by the suffix "TF" in Table \ref{sample-table}.
We evaluate model transferability with two metrics:
the zero-shot success rate and the final success rate.
The zero-shot success rate is calculated by directly 
evaluating the policy acquired in CSI-27/3/9 in 
larger-scale new environments (such as CSI-216/24/9).
We then continue to train the transferred policy in the new environment
and observe the success rate after training once again in the new environment.
The maximum evaluation success rate during this second training process is considered as the final success rate.
The transferring process is illustrated in Fig.\ref{fig:Transferability}.

HCGL has shown excellent performance in transferability.
Firstly,
HCGL has high zero-shot success rates.
The success rate only drops from 97\% to 77\% when the number of agents doubles (54 agents),
and to 75\% when the number of agents increases threefold (81 agents).
We further expand the size of the team 8 times (216 agents),
and the zero-shot success rate is still higher than 60\%.
Secondly,
transferred HCGL (HCGL-TF) can be trained to restore success rate again.
In all transferred tasks,
the final success rate of HCGL-TF is restored above 80\%.
% Furthermore,
% Fig.\ref{fig:Transferability} indicates that 
% the number of episodes required to restore the success rate in large-scale tasks is 
% significantly lower than training from scratches in 27 agents task.
Furthermore, as depicted in Figure 1, 
the number of episodes required to recover the success rate in large-scale tasks is 
significantly lower than training from scratch in the 27-agents task.

\textbf{Discussion.} The transferability simulations suggest that
difficult large-scale tasks can be solved by addressing less difficult tasks first,
then gradually increasing the number of agents and 
task difficulty until eventual challenges are solved.
The extensible feature of ECG allows us to utilize our HCGL model in various curriculum learning methods.

\subsection{Ablation Studies.}

\begin{figure}[!t]
    \centering
    \includegraphics[width=\linewidth]{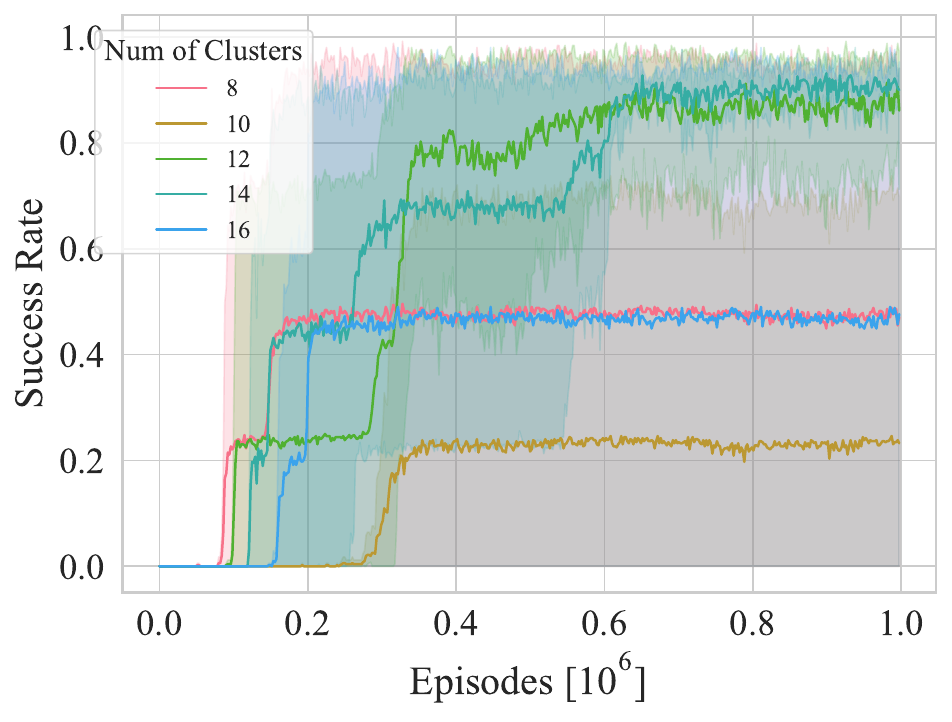}
    \caption{
        An ablation study on the influence of the number of clusters.
    }
    \label{fig:ablation-num-clusters}
\end{figure}

This experiment uses CSI-21/3/9 to investigate the influence of hyper-parameter settings in HCGL.

\subsubsection{The ablation study of clusters.}
First, $n_k=14$ clusters are set as the baseline.
To investigate the influence of $n_k$, 
we try multiple selections of $n_k$ in task CSI-21/3/9,
from $n_k=8$ to $n_k=16$.
As illustrated in Fig.\ref{fig:ablation-num-clusters},
the optimal number of clusters is between 12 to 14, which roughly matches the number of invaders plus the number of bases in CSI-21/3/9.

\subsubsection{The ablation study of primitive actions.}
Furthermore,
we test the influence of primitive actions.
The baseline model uses six primitive actions,
which move agents to $+x,-x,+y,-y,+z,-z$ respectively.
The 14-primitive model includes additional eight directions 
for $(+1,+1,+1),\ (+1,+1,-1),\ \dots$, etc.

\begin{figure}[!t] 
    \centering
    \includegraphics[width=\linewidth]{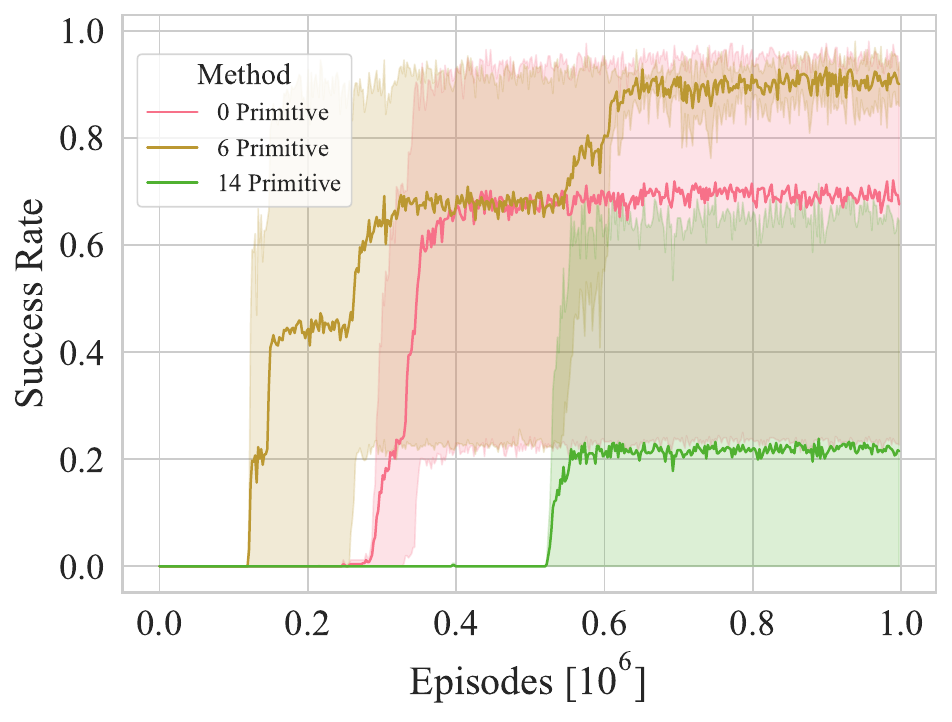}
    \caption{
        An ablation study on the influence of primitive actions.
    }
    \label{fig:ablation_primitive}
\end{figure}

The results displayed in Fig.\ref{fig:ablation_primitive} indicate
that the HCGL also works without any primitive actions,
but the average success rate declined significantly.
Additionally, 
a large number of primitive actions can also affect the performance.
Because increasing the operator action space can result in an increase 
in the required number of samples for policy exploration.

\textbf{Discussion}. 
Based on the simulation results presented in this section, selecting the optimal number of clusters and primitive actions is crucial for achieving optimal system performance.
The selection of an appropriate number of clusters depends on both the task at hand and the number of agents involved.
Additionally, HCGL exhibits superior performance when both cooperative and primitive actions are employed.

\begin{figure}[!t] 
    \centering
    \includegraphics[width=\linewidth]{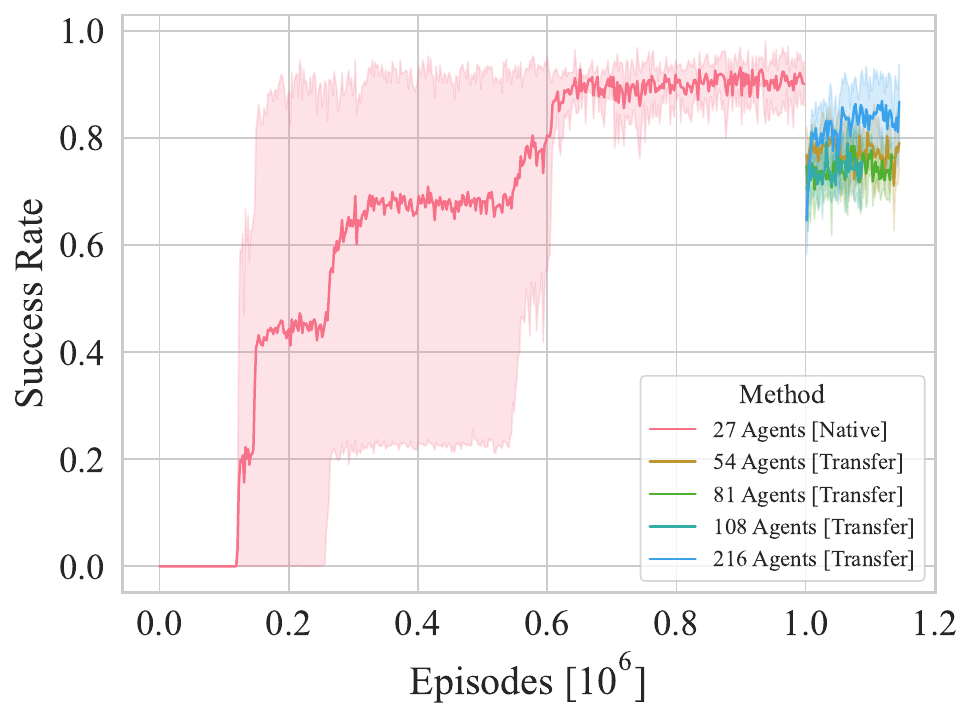}
    \caption{
        Testing transferability of HCGL. 
        The policy is firstly trained in task CSI-27/3/9,
        then the trained policy is transferred to CSI-54/6/9, CSI-108/12/9, etc.
    }
    \label{fig:Transferability}
\end{figure}

% \subsection{Expanding ECG for Larger-Scale Tasks.}

% \subsection{Expanability.}

% \subsection{Ablation Test of Target Nodes.}

% \subsection{Zero-Shot Scalability.}

% \subsection{Generalizing to Other Benchmarks.}

%
%
%
%

%% file: tex/06-conclusion.tex
\section{Conclusions.}

This paper proposes a novel MARL learning framework to 
address multi-agent problems that require sophisticated hierarchical collaboration.
Unlike classic MARL models,
we introduce a dynamic graph structure called the Extensible Cooperation Graph (ECG) 
to guide agents' cooperative behaviors.
ECG comprises three types of nodes: agents, clusters and targets.
Through dynamic edge adjustment,
ECG groups agents into clusters for carrying out 
either primitive individual actions or cooperative group actions.
To enable such adjustments in ECG,
we assign the authority to change edge topology to a group of graph operators,
who have a specialized action space for manipulating the ECG and 
neural policy networks to acquire ECG manipulation skills through reinforcement learning.

In our experiments,
we demonstrate that the ECG model not also has outstanding performance but also
exhibits both interpretability and transferability.
Furthermore,
its unique characteristic enables it to seamlessly incorporate cooperative knowledge from humans.